\documentclass[runningheads,orivec]{llncs}
\usepackage{multirow}
\usepackage{makecell}
\usepackage{booktabs}
\usepackage[T1]{fontenc}
\usepackage{amsmath}
\usepackage{amssymb}
\usepackage{xcolor}
\usepackage{url}
\usepackage{cite}

\usepackage{graphicx,verbatim}
\begin{document}
\title{Reducing Variability of Multiple Instance Learning Methods for Digital Pathology}
\titlerunning{Reducing Variability of MIL Methods} 



\author{Ali Mammadov \inst{1,2} \and Loïc Le Folgoc \inst{1} \and Guillaume Hocquet \inst{2} \and Pietro Gori \inst{1}}  
\authorrunning{A. Mammadov et al.}
\institute{LTCI, Télécom Paris, Institut Polytechnique de Paris, France \and
 Paris Saint-Joseph Hospital, France\\
\email{ali.mammadov@ip-paris.fr}}

\maketitle              
\begin{abstract}
Digital pathology has revolutionized the field by enabling the digitization of tissue samples into whole slide images (WSIs). However, the high resolution and large size of WSIs present significant challenges when it comes to applying Deep Learning models. As a solution, WSIs are often divided into smaller patches with a global label (\textit{i.e., diagnostic}) per slide, instead of a (too) costly pixel-wise annotation. By treating each slide as a bag of patches,  Multiple Instance Learning (MIL) methods have emerged as a suitable solution for WSI classification. A major drawback of MIL methods is their high variability in performance across different runs, which can reach up to 10-15 AUC points on the test set, making it difficult to compare different MIL methods reliably. This variability mainly comes from three factors: i) weight initialization, ii) batch (shuffling) ordering, iii) and learning rate. To address that, we introduce a 
\textit{Multi-Fidelity, Model Fusion} strategy for MIL methods.
We first train multiple models for a few epochs and average the most stable and promising ones based on validation scores. This approach can be applied to any existing MIL model to reduce performance variability. It also simplifies hyperparameter tuning and improves reproducibility while maintaining computational efficiency. We extensively validate our approach on WSI classification tasks using 2 different datasets, 3 initialization strategies and 5 MIL methods, for a total of more than 2000 experiments.   
\keywords{Multiple Instance Learning  \and Variadion Reduction \and Whole Slide Image Classification.}

\end{abstract}

\section{Introduction}

\begin{figure}[ht]
  {\includegraphics[width=\linewidth]{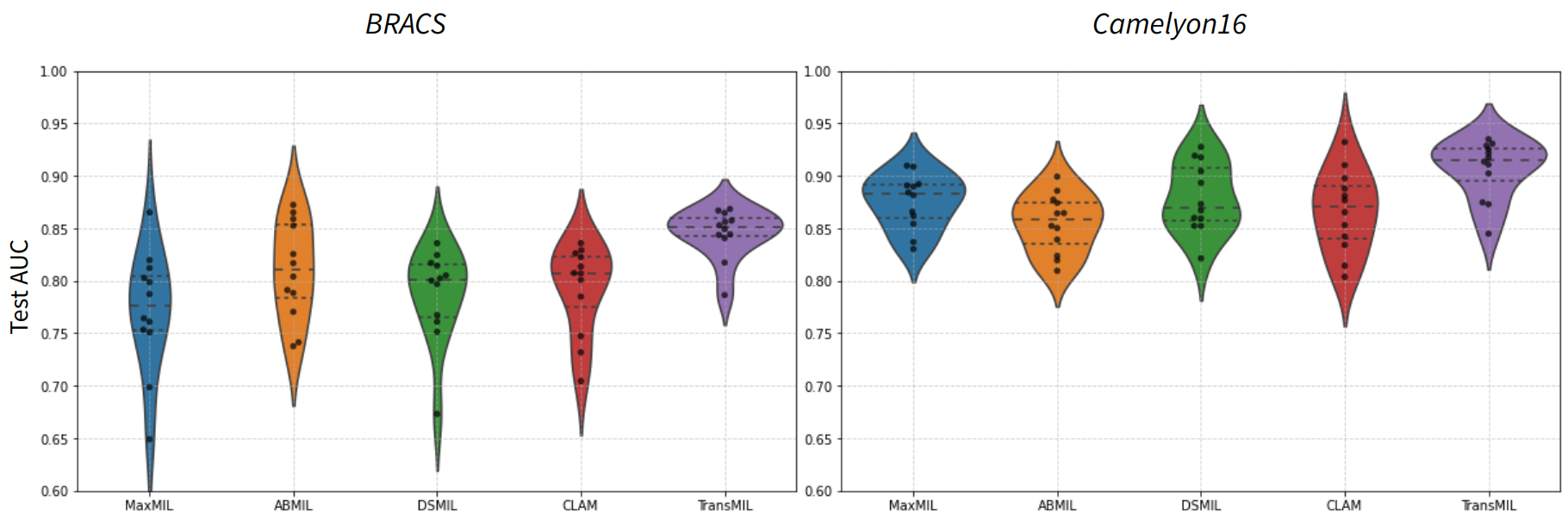}}
{\includegraphics[width=\linewidth]{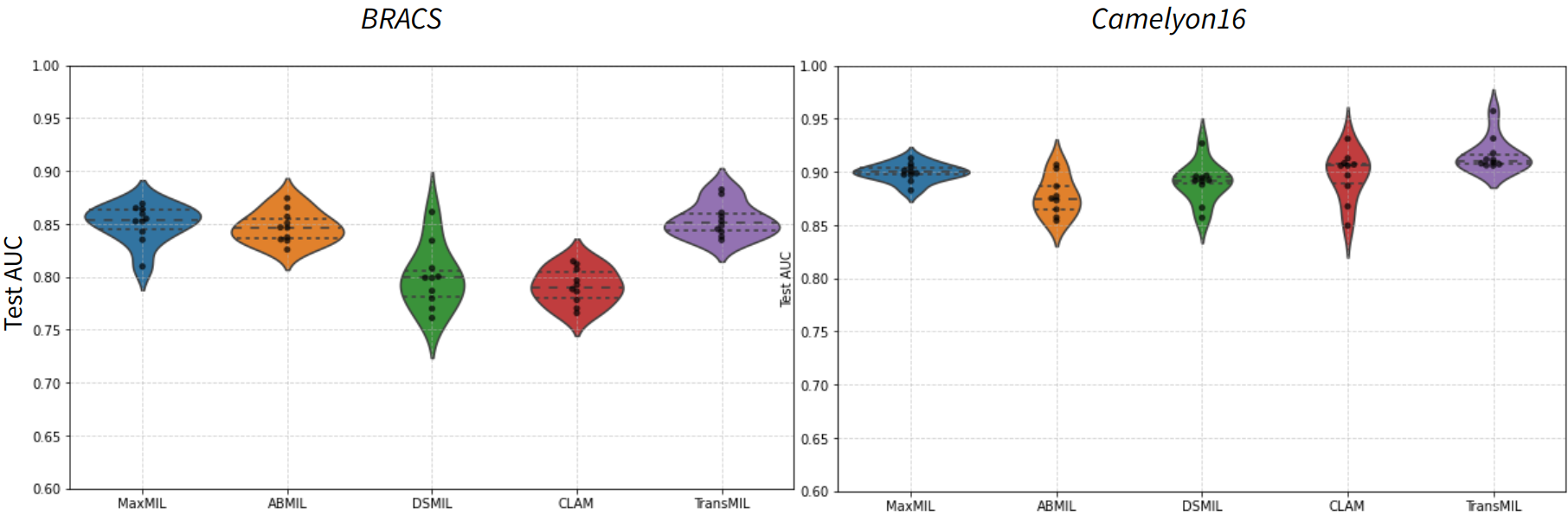}}
  { \caption{\textbf{Top}: Violin plots on 2 different datasets. Each dot represents the test AUC of a model trained with a random shuffle and initialization seed and with learning rate tuned on the validation set. \textbf{Bottom}: we apply the proposed method (using Soup for MaxMIL and ABMIL and Ties for DSMIL, CLAM and TransMIL) using the \textit{same} initialization seeds as in the Top figure and $M=10$, $K=5$ and $T=3$. The proposed method clearly reduces variability between different runs while preserving the top performance.}}
  \label{violin}
\end{figure}

Recent advances in Digital Pathology have made automated disease diagnosis using Deep Learning (DL) very popular. In these applications, a pathological slide is converted into a Whole Slide Image (WSI) in a pyramidal format, where each layer represents a different magnification. Because these images are very large, conventional DL methods are not practical. Instead, the Multiple-Instance Learning (MIL) framework is used for WSI classification.\\
In MIL, each slide is divided into small, non-overlapping patches using a sliding-window approach. These patches form a "bag" of instances. Unlike standard supervised learning, only the slide-level (bag-level) labels are available. This approach eliminates the need for expensive manual pixel-level annotations. A bag is labeled as negative if all patches are negative, and as positive if at least one patch is positive, which fits well with the fact that tumor regions often cover only part of the slide. First, semantically rich features are extracted from these patches using pre-trained encoders (from ImageNet or in a self-supervised way) or foundation models.
After feature extraction, either a patch-level classifier is trained and its scores are aggregated, or an aggregator is trained to create a slide-level representation that is used for the final prediction.\\
One ongoing challenge in deep learning is reproducibility. The performance of models often varies between runs, making it hard to compare models and tune hyperparameters. This problem appears in many DL fields \cite{gundersen2022sources,pineau2021improving}, such as natural language processing \cite{belz2021systematic}, generative adversarial networks \cite{lucic2018gans}, deep reinforcement learning \cite{henderson2018deep}, and image recognition \cite{bouthillier2019unreproducible}.\\
This issue also affects digital pathology for WSI classification with MIL. Recent works \cite{zhang2024attention,yang2024mambamil,jaume2024multistain,li2024dynamic,chan2023histopathology,lazard2023giga,chen2022scaling,zheng2022graph,wang2024dual} report high standard deviations between different runs of the same model.  
Furthermore, in all these works, the difference between the best performing method (usually the proposed one) and the second best performing one is very small, usually around 1-2 AUC, and much smaller than the variability of each method. This represents a significant problem for assessing whether a method actually outperforms the other methods or whether the reported differences are merely due to chance (also called "cherry picking").\\
In our experiments on two datasets with several MIL methods, we observed differences of up to 10--15 AUC points between runs. We found that this variation is mainly due to three factors: model initialization, the order of data presentation during training, and the way model weights are updated. We simplify these factors as:  the initialization seed, shuffle seed, and learning rate. Finding the perfect combination of these parameters is computationally expensive. 
Figure~\ref{violin}-Top shows the test AUC scores for 5 MIL methods on two different datasets (BRACS \cite{brancati_bracs_2022} and Camelyon \cite{ehteshami_bejnordi_diagnostic_2017}). For each MIL method, we tried 12 different combinations of shuffle and initialization seeds (black points), tuning the learning rate on the validation set. 

\noindent \textbf{Related Works} One of the earliest approaches to build more robust models is ensemble modeling. The idea is straightforward: instead of training a single model, multiple models are trained, and during inference, their predictions are averaged. Another simple solution is to just pick the model with the best Validation score. However, these approaches have a major drawback: high computational cost, as they require fully training several models. To address this, Wortsman et al. \cite{wortsman2022model} introduced Model Soups (or Model Averaging), a method that averages the weights of multiple models, trained from the same parameter initialization, allowing a single forward pass while maintaining the benefits of ensembling. This approach improves performance and robustness without increasing inference costs. However, it may arise another issue when averaging weights with opposite signs, as they can cancel each other out, leading to inactive neurons. \textit{TIES-Merging} \cite{yadav2024ties} solves this by averaging only the weights with matching signs and setting small conflicting values to zero.

\noindent \textbf{Contributions} Inspired by model averaging \cite{wortsman2022model,yadav2024ties} and multi-fidelity hyperparameter optimization \cite{giselle2023multifidelity,egele2023epoch}, we propose a new method to reduce performance variability in MIL-based whole slide image classification. 
In our work, we use the idea of \emph{Model Soups} and \textit{TIES-Merging}, where we average the weights of the best models. For choosing the best models we follow the idea of multi-fidelity hyperparameter optimization. Instead of fully training each model, we train it for a few epochs to quickly estimate its performance. 
By combining these two approaches, our goal is to smooth out differences caused by random initialization, data shuffling, and training updates, thus reducing performance variability. In our method, we first train $M$ models for $K$ epochs (usually $M=10$ and $K=5$), then select the top $T$ models (usually $T=3$) based on early validation AUC scores and eventually average their weights. We show that this simple procedure reduces the performance variability thus increasing reproducibility and trustworthiness. Meanwhile, our method keeps the computational burden at a reasonable rate, increasing the total number of training epochs of only $K*M=50$, which usually represents 25\% or 50\% of the total number of training epochs.
Additionally, this method is generic and can be applied to any existing MIL method. 

\section{Method}
\textbf{MIL Formulation.} Each slide is modeled as a labeled bag containing unlabeled patches. Consider dataset $\mathcal{S}$ containing \(N\) slides, represented as S=\(X_1, X_2, \ldots X_N\) associated with labels denoted by \(Y = \{y_1, y_2, \ldots, y_N\}\). Each individual slide \(X_i\) is composed of a collection of patches, denoted as \(X_i=\{x_{i,1}, x_{i,2}, \ldots, x_{i,P_{i}}\}\), extracted exclusively from the foreground tissue regions of the slide where the value of $P_{i}$ varies based on the size of the slide. Note that there are no labels for patches($x_{i,j}$), only slide-level labels are provided, therefore this is considered as a weak supervision. The WSI classification pipeline is structured into multiple phases. The initial phase is pre-training, during which the backbone model $\mathbf{f_\phi}$ is pre-trained on the patches of the training slides with the given self-supervised learning method.
Then, for every slide $i$, features are extracted from patches $j$, assembled within each bag, and used as input for the MIL aggregator network $\mathbf{g_{\theta_g}}$. This network aggregates the features to generate a bag representation of the slide $i$, which is then forwarded to the classifier $\mathbf{c_{\theta_c}}$ for predicting the class based on the task. It can be formulated as:
\begin{equation}
  h_{i,j} = \mathbf{f_\phi}(x_{i,j}); \quad H_i = \mathbf{g_{\theta_g}}(h_{i,1}, h_{i,2}, \ldots, h_{i,P}); \quad C_i = \mathbf{c_{\theta_c}}(H_i) 
\end{equation} 
In this work, we ignore the variability of the encoder $f_\phi$, considering it already pre-trained and frozen, focusing only on the other two networks, namely $g_{\theta_g}$ and $c_{\theta_c}$, which present a variability that depends on their gradient-based optimization process. The values of the final parameters $\theta=\{\theta_g,\theta_c \}$ (i.e., at the end of the training) depend on the initialization and on the optimization process, whose most important hyper-parameters are: initialization seed, shuffling seed and learning rate. By changing one of them, results may drastically vary.

\begin{figure}[ht!]
    \centering
    \includegraphics[width=\linewidth]{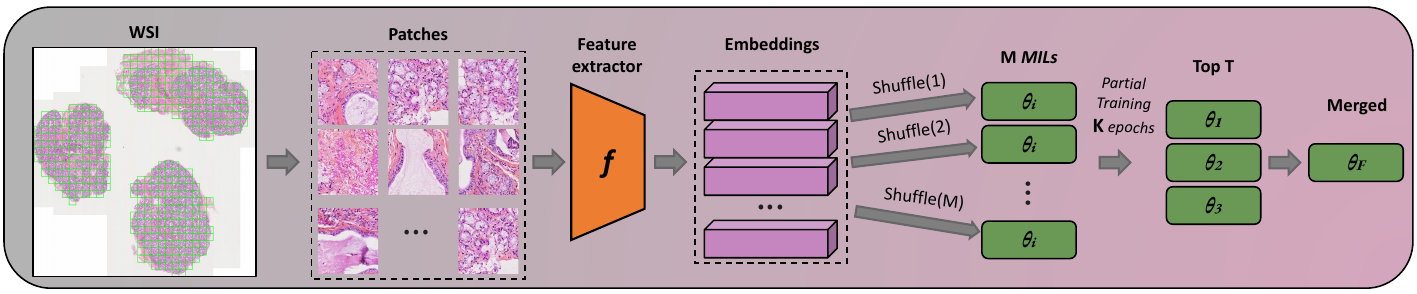}
    \caption{ Pipeline of the proposed Multi-Fidelity Model Averaging method for Whole Slide Image Classification. }
    \label{fig:pipeline}
\end{figure} 

\noindent \textbf{Model Overview} In our work, we propose combining Multi-Fidelity with Model Averaging to mitigate variability and increase robustness (see Fig.\ref{fig:pipeline}). Each slide is first cut into patches, whose features are extracted using $f_\phi$. Then, $M$ \textbf{identical} models are initialized with the same initialization seed, and each model is trained with a different and random shuffling seed for a small number of $K$ epochs. The learning rate can be tuned using the validation set or chosen based on prior knowledge from existing literature, without further tuning. After the partial training, the top $T$ (e.g., 3) models are selected based on their validation AUC scores. Next, the weights of these selected models are aggregated, and the resulting, combined model is fully trained. 

\noindent \textbf{Model aggregation}
We propose using two simple merging methods: uniform \textit{SOUP} \cite{wortsman2022model} and \textit{TIES}-Merging \cite{yadav2024ties}. Let $\theta_1, \theta_2, \dots, \theta_M$ be the weights of the $M$ partially trained models. Uniform SOUP is simply defined as the average across models: $\theta_{\text{uniform}} = \frac{1}{M}\sum_{i=1}^{M}\theta_i$. 
TIES-Merging works by first calculating the difference between each partially trained model and the initial model. It then trims these differences to keep only the most important changes, discarding the small ones. Next, for each parameter, it chooses the dominant sign among the models and it averages only the values that agree with the dominant sign. 

\section{Results and Discussion}\label{exp}

\subsection{Implementation and Data information}
\textbf{Datasets and Data Splits.} We conduct experiments on two datasets. Camelyon16 \cite{ehteshami_bejnordi_diagnostic_2017} is a 2-class dataset for the detection of metastases in breast cancer. It comprises 400 slides, with 239 normal tissue slides and 160 tumor slides. BReAst Carcinoma Subtyping (BRACS) \cite{brancati_bracs_2022} is a 3-class imbalanced dataset for breast carcinoma subtyping, containing 547 whole-slide images (WSIs): 265 benign tumor cases, 89 atypical tumor cases, and 193 malignant tumor cases. For both datasets, we use the official data splits. \\
\textbf{Evaluation Metric.} Our main evaluation metric is the AUC score, which is resilient to class imbalance effects. We select the best-performing models on the validation set and report their AUC scores on the test sets.\\
\textbf{Pre-processing.} Following CLAM's pre-processing pipeline \cite{lu_data-efficient_2021}, we cut all WSIs into 256×256 non-overlapping patches extracted solely from foreground tissue regions at x10 magnification. This results in approximately 0.6 million patches for Camelyon16 and 1.4 million patches for BRACS. \\
\textbf{Feature Extraction.} We extract features using self-supervised learning-based pre-trained backbones. For Camelyon16, we pre-train a ResNet18 (11.7M parameters) with Barlow-Twins \cite{zbontar_barlow_2021}. For BRACS, we also use ResNet18 but pre-train it with DINO \cite{caron_emerging_2021}, to ensure that variations in results are not dependent on the pre-training method and since these two methods demonstrated state-of-the-art results \cite{kang_benchmarking_2023}. All pre-training is conducted with the \textit{solo-learn} library \cite{costa_solo-learn_2022} for 200 epochs, with SSL hyper-parameters kept as in the original papers.\\
\textbf{Training and Evaluation.} We adopt DSMIL’s code \cite{li_dual-stream_2021} as the base for our training and evaluation pipeline, and we include five state-of-the-art (SOTA) MIL methods in our study: MaxMIL (baseline), ABMIL \cite{ilse_attention-based_2018}, DSMIL \cite{li_dual-stream_2021}, CLAM \cite{lu_data-efficient_2021}, and TransMIL \cite{shao_transmil_2021}. We use a cosine annealing scheduler, the Adam optimizer with a weight decay of 0.00001, and a batch size equal to one slide (i.e., one bag). For further details on the hyper-parameters, please refer to the released code \url{https://anonymous.4open.science/r/mil_merging-EE39/}. 

\subsection{Experiments}
\noindent \textbf{Variability Analysis} We evaluate our methods (Soup and Ties) on two datasets (Camelyon16 and BRACS), across five MIL models. We compare them with four other methods: \textit{Baseline}, \textit{LR tuned}, \textit{Ensemble} and \textit{Best on Val}. Each method is evaluated using ten different initialization seeds, thus obtaining ten different AUC scores on the test sets. The variability of the performances is evaluated using four metrics: minimum AUC, maximum AUC, mean AUC, and standard deviation.\\
Here, we give a brief description of each method, given an initialization seed:\\
\textit{Baseline} represents a single model that is trained for 100 epochs with random shuffling.\\
For \textit{LR tuned} we perform a supplementary grid search over 6 learning rates. Each model is trained for 100 epochs, and we pick the best one on the validation set. The total number of training epochs is thus 600.\\ \textit{Soup3} and \textit{Ties3} are our proposed methods with parameters \(M=10\), \(K=5\) and \(T=3\), which require \(M \times K + 100 = 150\) epochs of training.\\ \textit{Ensemble} is an average of predictions from 10 fully trained models (each trained for 100 epochs with a random shuffling seed), resulting in 1000 total epochs.\\ In \textit{Best on VAL}, we select 
the model with the highest validation score among the 10 fully trained models and report its test performance (also 1000 total epochs).\\
\noindent \textbf{Ablation Study.} We have conducted two ablation studies. In first one, we evaluate the influence of the hyper-parameters $K$, $T$, and initialization type using the MaxMIL method and $M=10$. We compare three highly-used and well-known initialization strategies: i) \textit{Uniform} initialization, where the initial weights are drawn from a uniform distribution with mean $\mu=0$ and standard deviation $\sigma=1$, ii) \textit{Xavier}, weights are sampled from a normal distribution with $\mu=0$ and $\sigma=\sqrt{\frac{2}{fan_{in}+fan_{out}}}$, where $fan_{in}$/$fan_{out}$ are number of input/output signals, and iii) \textit{Switch} initialization is done by drawing initial weights from a truncated normal distribution with $\mu=0$ and $\sigma=\sqrt{s/n}$, where s is a scale hyper-parameter and n is the number of input units in the weight tensor. In the second study, we investigate the effect of the number of models to merge ($T$) on the performance of \textit{Soup} and \textit{Ties} methods across both datasets and all MILs, using $M=10$ and $K=100$ (thus each model is trained for 100 epochs before aggregation). Here, for each MIL, we change the value of T from 2 to 10 reporting the average AUC score on the test set of 10 runs. 
\begin{figure}[htbp]
  {\includegraphics[width=\linewidth]{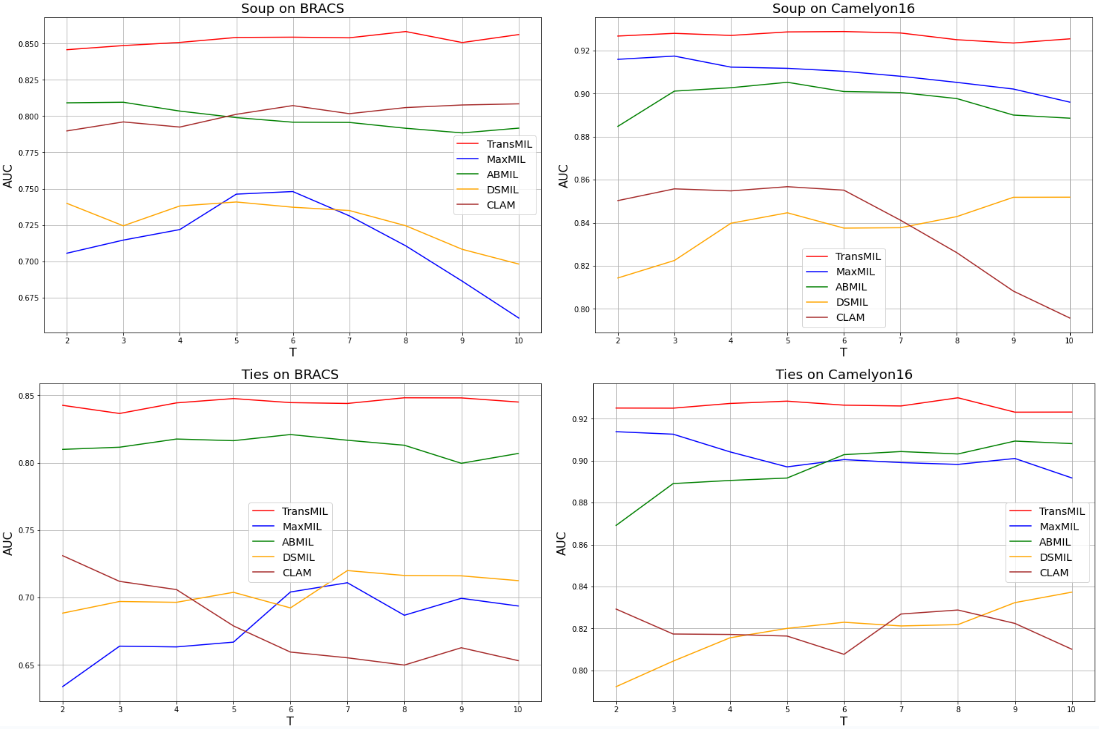}}
  \caption{ Ablation Study on the effect of the T to the average AUC score across 5 MILs on Camelyon16 and BRACS datasets }
  \label{fig:abl_T}
\end{figure}
\subsection{Discussion} The proposed methods achieve better results in Table \ref{std} than \textbf{Baseline} and \textbf{LR tuned} based on all metrics across all MILs and on both datasets. Only for BRACS, the Mean and Max of CLAM from \textit{Baseline} and \textit{LR tuned} is slightly better (0.1 AUC point), but it is important to consider that \textit{LR tuned} requires 4 times more training epochs.   
Furthermore, our proposed methods obtain more stable results across all MIL methods and datasets, having the smallest \textbf{STD}. It's also important to notice that the proposed methods have a similar or better performance than \textbf{Ensemble} and \textbf{Best on VAL} models,  but they require almost 7 times less training epochs and they are 10 times faster at inference. Eventually, it's worth mentioning that  \textit{Soup3/Ties3} improve the minimum results, which means that multi-fidelity based training helps to avoid local minima during training, and the best-performing methods for both dataset are our proposed methods with a maximum AUC of 95.7 points for Camelyon16 and 88.2 points for BRACS. 

From the Ablation study (Table \ref{ablation}), we can see that results using $K=10$ are slightly better than $K=5$ or $K=3$. However, this comes at a cost of increasing the number of training epochs. To keep the computational burden low, while preserving a good performance, we chose the combination $K=5$ and $T=3$, which gives almost always the best or second best performances on the Val and Test set (using $M=10$). This means that aggregating 3 models out of 10 seems to be a good compromise between stability and performance. This is why we chose $T=3$ in Table \ref{std}. 

The ablation study presented in Fig. \ref{fig:abl_T} shows that increasing the number of models to merge does not significantly improve the average performance. Best results are obtained, on average, with a value between $T=3$ and $T=5$. $T=3$ seems thus a good choice.

\begin{table*}[ht]
    \centering
    \caption{Variability analysis. Each method is repeated 10 times using 10 different initialization seeds. The 10 AUC on the test set are then used to compute the variability measures (Min, Max, Mean and STD). For each method and a given initialization seed, we report the number of models $M$ (i.e., different shuffling seeds), the number of epochs $K$ of initial training and the total number of training epochs $Ep$.}
    \resizebox{\textwidth}{!}{
    \begin{tabular}{cccc |cccc|cccc|cccc|cccc|cccc}
    \toprule
    \multicolumn{24}{c}{\textbf{Camelyon16 Dataset}} \\
    \toprule
    \multirow{2}{*}{\textbf{Method}}& 
    \multirow{2}{*}
    {\textbf{M}} &  
    \multirow{2}{*}
    {\textbf{K}} &  
    \multirow{2}{*}{\textbf{Ep}} &  \multicolumn{4}{c}{\textbf{MaxMIL}} &  \multicolumn{4}{c}{\textbf{ABMIL}} &  \multicolumn{4}{c}{\textbf{DSMIL}} &  \multicolumn{4}{c}{\textbf{CLAM}} &  \multicolumn{4}{c}{\textbf{TransMIL}} \\
    \cmidrule(lr){5-24}  
     &  & &  & Min & Max & Mean & STD & Min & Max & Mean & STD & Min & Max & Mean & STD & Min & Max & Mean & STD & Min & Max & Mean & STD \\
    \cmidrule(lr){1-24}  
    \textit{Baseline}   & 1 & -  & 100 & 75.8 & 90.7 & 88.0 & 4.3 & 76.4 & 90.7 & 83.7 & 3.9 & 79.7 & 91.6 & 86.2 & 3.6 & 70.2 & 92.7 & 83.8 & 5.9 & 85.7 & 92.6 & 90.3 & 2.0 \\
    \textit{LR tuned}  & 6 & - & 600 & 83.0 & 91.0 & 87.6 & 2.5 & 81.0 & 89.9 & 85.5 & 2.7 & 82.1 & 92.8 & 87.9 & 3.2 & 80.4 & 93.2 & 86.7 & 3.7 & 84.5 & 93.5 & 90.6 & 2.7 \\
    \cmidrule(lr){1-24}  
    \textit{Soup3 (ours)}    & 10 & 5  & 150 & 88.3 & 91.3 & 89.9 & 0.8 & 85.4 & 90.7 & 87.8 & 1.7 & 84.3 & 91.8 & 87.8 & 2.6 & 84.8 & 93.5 & 89.5 & 2.6 & 84.3 & 88.4 & 86.4 & 1.4 \\
    \textit{Ties3 (ours)}   & 10 & 5     & 150 & 86.3 & 91.7 & 89.8 & 1.6 & 79.0 & 91.5 & 85.5 & 3.6 & 85.7 & 92.7 & 89.0 & 1.8 & 84.9 & 93.1 & 89.7 & 2.2 & 90.6 & 95.7 & 91.7 & 1.5 \\
    \cmidrule(lr){1-24}  
    \textit{Ensemble} & 10 & -    & 1000 & 88.0 & 92.7 & 90.3 & 1.3 & 86.6 & 93.2 & 91.1 & 2.1 & 79.1 & 88.4 & 83.9 & 3.1 & 84.9 & 94.3 & 87.8 & 2.7 & 92.0 & 94.9 & 93.5 & 0.9 \\
    \textit{Best on VAL} & 10 & -   & 1000 & 88.3 & 92.9 & 91.0 & 1.4 & 79.9 & 92.2 & 85.6 & 4.2 & 82.6 & 92.8 & 88.5 & 3.5 & 77.0 & 89.3 & 83.5 & 4.2 & 89.1 & 94.6 & 92.2 & 1.6 \\
    \toprule
    \toprule
    \multicolumn{24}{c}{\textbf{BRACS Dataset}} \\
    \toprule
    \textit{Baseline}  & 1 & -   & 100 & 50.8 & 74.3 & 58.7 & 7.3 & 73.5 & 83.3 & 78.4 & 3.1 & 66.1 & 81.3 & 73.8 & 4.7 & 71.2 & 83.5 & 79.6 & 3.1 & 79.7 & 86.4 & 83.1 & 2.0 \\
    \textit{LR tuned}  & 6 & -   & 600 & 64.9 & 86.5 & 77.2 & 5.5 & 73.8 & 87.2 & 81.0 & 4.4 & 67.3 & 83.6 & 78.7 & 4.2 & 70.4 & 83.6 & 79.3 & 4.1 & 78.6 & 86.8 & 84.6 & 2.2 \\
    \cmidrule(lr){1-24}  
    \textit{Soup3(ours)}   & 10 & 5     & 150 & 81.0 & 86.9 & 85.0 & 1.7 & 81.3 & 86.6 & 84.6 & 1.5 & 76.1 & 86.1 & 80.0 & 2.8 & 73.3 & 83.5 & 79.5 & 2.8 & 83.5 & 88.2 & 85.4 & 1.5 \\
    \textit{Ties3(ours)}    & 10 & 5    & 150 & 81.0 & 86.9 & 84.8 & 2.0 & 82.6 & 87.4 & 84.7 & 1.4 & 74.9 & 81.4 & 78.7 & 1.8 & 76.6 & 81.5 & 79.1 & 1.6 & 82.1 & 87.5 & 85.0 & 1.5 \\
    \cmidrule(lr){1-24}  
    \textit{Ensemble}   & 10 & -  & 1000 & 65.6 & 84.1 & 76.4 & 5.8 & 78.5 & 83.2 & 81.1 & 1.6 & 71.6 & 84.9 & 77.4 & 3.7 & 77.9 & 82.0 & 80.2 & 1.3 & 82.9 & 87.3 & 85.8 & 1.2 \\
    \textit{Best on VAL}   & 10 & -      & 1000 & 70.7 & 86.6 & 79.1 & 5.2 & 76.4 & 85.2 & 79.8 & 2.5 & 71.3 & 80.8 & 76.2 & 2.7 & 74.0 & 82.1 & 78.2 & 2.7 & 80.0 & 86.6 & 83.5 & 1.8 \\
    \toprule
    \toprule
    \end{tabular}} 
    \label{std}
\end{table*} 

\begin{table*}[ht]
    \centering
    \caption{Ablation study on: 1) number of epochs $K$, 2) number of aggregated models $T$ and 3) initialization type using MaxMIL and $M=10$.}
    \resizebox{\textwidth}{!}{
    \begin{tabular}{ccccccc|cccccc|cccccc}
    \toprule
    \multirow{3}{*}{\textbf{Method}} & \multicolumn{2}{c}{\textbf{Uniform}} &  \multicolumn{2}{c}{\textbf{Xavier}} &  \multicolumn{2}{c}{\textbf{Switch}}  & \multicolumn{2}{c}{\textbf{Uniform}} &  \multicolumn{2}{c}{\textbf{Xavier}} &  \multicolumn{2}{c}{\textbf{Switch}}  & \multicolumn{2}{c}{\textbf{Uniform}} &  \multicolumn{2}{c}{\textbf{Xavier}} &  \multicolumn{2}{c}{\textbf{Switch}} \\
    \cmidrule(lr){2-19}  
    & \multicolumn{6}{c}{\textit{K = 3 epochs}}  & \multicolumn{6}{c}{\textit{K = 5 epochs}}   & \multicolumn{6}{c}{\textit{K = 10 epochs}} \\
    \cmidrule(lr){2-19}  
    & Val & Test & Val & Test & Val & Test & Val & Test & Val & Test & Val & Test & Val & Test & Val & Test & Val & Test\\
    \cmidrule(lr){1-19}  
    \textit{Soup3 (T=3)}        & 98.2 & 89.6 & 98.6 & 89.1 & 97.6 & 89.1 & 98.4 & 92.5 & 98.2 & 91.8 & 98.0 & 89.8 & 98.6 & 93.3 & 98.8 & 91.7 & 99.6 & 91.7 \\
    \textit{Soup5 (T=5)}        & 98.0 & 89.0 & 98.4 & 88.8 & 97.6 & 90.3 & 98.0 & 89.5 & 97.8 & 89.8 & 96.8 & 89.6 & 98.0 & 90.0 & 97.6 & 89.3 & 98.4 & 89.3 \\
    \textit{Soup (T=10)}         & 98.0 & 91.5 & 99.8 & 92.3 & 98.6 & 88.9 & 98.0 & 89.2 & 99.8 & 91.7 & 99.0 & 90.4 & 98.6 & 93.7 & 98.6 & 91.6 & 98.8 & 90.7 \\
    \cmidrule(lr){1-19}   
    \toprule
    \end{tabular}} 
    \label{ablation}
\end{table*} 
\section{Conclusion}
MIL methods suffer from high variability in performance across different runs, which can hamper reproducibility and trustworthiness when comparing different methods. To address this issue, we introduced a simple strategy based on model averaging and multi-fidelity optimization. Our experiments demonstrated that the proposed method reduces performance variability across runs while preserving top performance and maintaining a sustainable computational burden.
\newline

\noindent \textbf{Acknowledgments.} This paper has been supported by the French National Research Agency (ANR-20-THIA-0012) and by the Hi!PARIS Center on Data Analytics and Artificial Intelligence. Furthermore, this work was performed using HPC resources from GENCI-IDRIS (Grant 2023-AD011013982R1).
%
%
%
\newpage
\bibliographystyle{splncs04}
\bibliography{mybibliography}

\end{document}